\documentclass{article}
\usepackage[utf8]{inputenc}
\usepackage[english]{babel}
\usepackage{multirow}
\usepackage{lipsum}
\usepackage{authblk}
\usepackage[top=3cm, bottom=3cm, left=4cm, right=4cm]{geometry}
\usepackage{fancyhdr}
\usepackage{caption}

\usepackage{biblatex}
\title{Multi-Agent Inverse Reinforcement Learning: \linebreak
Suboptimal Demonstrations and Alternative Solution Concepts}

\author{Sage Bergerson}
\affil{Stanford Existential Risk Initiative \linebreak
{\footnotesize Stanford University, 
Stanford, CA 94301\linebreak
sab9835@nyu.edu}}

\begin{document}
\maketitle

\begin{abstract}
Multi-agent inverse reinforcement learning (MIRL) can be used to learn reward functions from agents in social environments. To model realistic social dynamics and outcomes, MIRL methods must account for suboptimal human reasoning and behavior. Traditional formalisms of game theory provide computationally tractable behavioral models, but assume agents have unrealistic cognitive capabilities. The objective of this research is to identify and compare mechanisms used in MIRL methods to a) handle the presence of noise, biases and heuristics in agent decision making and b) model realistic equilibrium solution concepts based on suboptimal behavior. MIRL research was systematically reviewed to identify solution attempts for these challenges. The methods and results of these studies were analyzed and compared based on factors including performance accuracy, efficiency, and descriptive quality. We found that the primary methods for handling noise, biases and heuristics in MIRL were extensions of Maximum Entropy (MaxEnt) IRL to multi-agent settings. We also found that traditional Nash equilibrium (NE) solution concepts were ill-suited to model human behavior, and more success could be found with generalization of the NE. These solutions include the correlated equilibrium, logistic stochastic best response equilibrium and entropy regularized mean field NE. Methods which use recursive reasoning or updating also perform well, including the feedback NE and archive multi-agent adversarial IRL. Success in modeling specific biases and heuristics in single-agent IRL and promising results using a Theory of Mind approach in MIRL imply that modeling specific biases and heuristics in MIRL may be useful. Flexibility and unbiased inference in the identified alternative solution concepts suggest that a solution concept which has both recursive and generalized characteristics may perform well at modeling realistic social interactions.
\end{abstract}

\section{Introduction}

Inverse reinforcement learning (IRL) is the problem of learning the reward function being optimized for by an observed agent acting in a Markov decision process (MDP). The learned reward function may then be used in inner-loop reinforcement learning (RL) by the autonomous system to develop a policy which directs the system’s behavior \cite{articleNR}. IRL may prove useful for adopting skilled or desired behavior from natural systems with human agents. 

To learn accurate reward functions such that the RL component of IRL systems performs the desired behavior, IRL methods must be able to make inferences based on realistic demonstrations by agents. The conceptual and technical challenges of moving from simple, tractable representations of behavior to those more general (and thus computationally complex) comprise a large portion of current IRL research. 

The initial conceptualization of IRL, and the focus of much of the subsequent research in this area, has been on single-agent dynamics, i.e. learning a reward function based on observations of a single agent. However, IRL can also be generalized to multi-agent dynamics through multi-agent IRL (MIRL). 

MIRL methods can take several approaches to reward learning, such as modeling natural swarm systems \cite{10.5555/3091125.3091320}, collective agent populations \cite{yang2018learning, chen2021agentlevel}, or multiple individual agents in a shared environment \cite{Reddy2012InverseRL, Lin2018MultiagentIR}. MIRL methods may be applied to both cooperative \cite{9308491} and competitive \cite{zhang2020noncooperative, wang2018competitive} dynamics. In addition to the capabilities achieved by single-agent IRL, MIRL may also be able to coordinate systems of agents to find maximally beneficial social outcomes. However, this comes with the added challenge of accounting for the influence of multi-agent dynamics on behavior. Due to the potential added benefits from the use of MIRL methods, these challenges are worth addressing. 

To do so, this paper analyzes, compares and contrasts techniques used in recent MIRL research to innovate along two dimensions: handling suboptimal human planners and implementing flexible alternatives to traditional solution concepts for multi-agent interactions.

A key promise of IRL is that of reliably and efficiently uncovering rewards of humans in the context of human-computer interactions \cite{articleNR}. However, biased human planners are subject to disproportionate inclinations to prefer one alternative over another and rely on heuristics, mental shortcuts used to make efficient decisions. Handling suboptimal human planners is also a challenge for single-agent IRL, but multi-agent dynamics add additional considerations, such as the effect of suboptimality on interactions and the ability of agents to account for bias in others. In contradistinction to traditional game theoretic formalisms of representing human behavior with high computational abilities and information access, a suboptimal human planner can only possess the mental abilities and information characteristic of a human decision-maker \cite{Simon1955ABM}. 

Similarly, traditional solution concepts concepts generally only apply in the context of unrealistic rationality for human decision makers \cite{10.1007/978-3-642-25280-8_1}. Solution concepts are game theoretic formalisms modeling the dynamics of social interactions which predict how a game will be played. A commonly used solution concept is the Nash equilibrium (NE), which provides a tractable formalism for dynamic social behavior \cite{nash1950equilibrium}. While the NE is useful, it suffers from several practical limitations: restriction to settings with a finite number of moves and players, the possibility of several outcomes of equal value with no mechanism for ultimate selection, and assumptions of agents' mutual knowledge and unwavering rationality. Therefore, exploring more general and flexible solution concepts is necessary. 

Our research highlight that extensions of the maximum entropy (MaxEnt) framework to multi-agent scenarios can accommodate forms of noise (random variability) and bias (systematic variability) in decision making. Modeling specific biases and heuristics has shown to be useful in single-agent IRL and promising results from a multi-agent formalism using a Theory of Mind (ToM) approach suggest modeling specific cognitive characteristics would be useful for multi-agent systems as well. Our research also highlights that more generalized solution concepts than the NE, including the correlated equilibrium (CE), and logistic stochastic best response equilibrium (LSBRE), and entropy regularized mean field Nash equilibrium (ERMFNE) as well as solution concepts that employ recursive reasoning or updating based on cooperative trajectory identification, may be more flexible and accommodating of realistic behavior. 

\section{Realistic behavior: Noise, biases, heuristics} \label{Model}

To make accurate inferences, it is important for MIRL methods to accommodate naturalistic human agent behavior. To do this, it is essential for methods to be flexible in the presence of noise, biases and heuristics in decision making processes.

\subsection{Assumption of Rationality}

Early methods of IRL frequently assume that observed expert demonstrations are optimal \cite{articleNR}. An agent demonstration is considered expert behavior when the agent can reach solutions through the use of computations or information unattainable or unavailable to a novice \cite{simonInvariantsHumanBehavior1990}. This assumption of value-based optimality can be equated with ideal practical rationality, which finds its origins in economics, rational choice theory, and decision theory \cite{Edwards1954TheTO, 10.2307/j.ctt1r2gkx}. Ideal practical rationality refers to the rule that, given a set of competing alternatives, the optimal choice for an agent tasked with making the right decision is that which maximizes expected utility \cite{Abel2019ConceptsIB}. This forms the basis of expected utility theory (EUT) and provides a compact theory through which empirical predictions of behavior can be made from a relatively sparse problem model — simply a description of the agent’s objectives and constraints. Ideal practical rationality is an appealing choice for modeling human behavior through machine learning models due to its use of a utility function in an optimization-based formula \cite{Levin2004IntroductionTC}. 

However, EUT has been shown to fail in some cases, as demonstrated by the Allais paradox, revealing inconsistencies between the actual choices observed in humans and those anticipated by EUT \cite{10.2307/1907921}. The paradox arises when comparing participants’ choices in two experiments, each of which involves a choice between two gambles. The payoffs of the gambles are described in Table 1. Studies show that when choosing between 1A and 1B, most participants choose 1A. Likewise, when choosing between 2A and 2B, most participants choose 2B \cite{10.1257/jep.1.1.121}. However, according to EUT, a rational person should choose either 1A and 2A or 1B and 2B, due to the relative equality of payoffs for these scenarios. 

\begin{figure}
    \centering
\begin{tabular}{|l|l|l|l|l|l|l|l|}
\hline
\multicolumn{4}{|l|}{Experiment 1}                                                   & \multicolumn{4}{l|}{Experiment 2}                                                                 \\ \hline
\multicolumn{2}{|l|}{Gamble 1A}                     & \multicolumn{2}{l|}{Gamble 1B} & \multicolumn{2}{l|}{Gamble 2A}                     & \multicolumn{2}{l|}{Gamble 2B}               \\ \hline
Prize                      & Prob.                  & Prize            & Prob.       & Prize                      & Prob.                 & Prize                & Prob.                 \\ \hline
\multirow{3}{*}{\$1 mill.} & \multirow{3}{*}{100\%} & \$1 mill.        & 89\%        & \$0                        & 89\%                  & \multirow{2}{*}{\$0} & \multirow{2}{*}{90\%} \\ \cline{3-6}
                           &                        & \$0              & 1\%         & \multirow{2}{*}{\$1 mill.} & \multirow{2}{*}{11\%} &                      &                       \\ \cline{3-4} \cline{7-8} 
                           &                        & \$5 mill.        & 10\%        &                            &                       & \$5 mill.            & 10\%                  \\ \hline
\end{tabular}
    \caption*{Table 1: Payoffs for the two experiments describing the Allais paradox. According to EUT, a rational actor would select either Gamble 1A and 2A or Gamble 1B and 2B. However, experiments find that most participants will select Gamble 1A in Experiment 1 and Gamble 2B in Experiment 2.}
    \label{fig:my_label0}
\end{figure}

In fact, human choices often do not display the stability and coherence of ideal practical rationality as defined by EUT. Two types of deviations commonly occur in natural human behavior: noise (random variation) and biases and heuristics (systematic variation). The factors behind seemingly irrational human decision making (i.e. contextual and symbolic complexity, and the influence of meaning-based and ethical preferences) may be indispensable to the human decision making process \cite{stanovichWhyHumansAre2013}. 

Methods for handling both deviations have been explored in MIRL research. The principle of maximum entropy has been used to produce superior results in both cases. Additional techniques, including a ToM approach to model bounded intelligence, have been used to address biases and heuristics in multi-agent settings.

\subsection{Noise}

Noise refers to chance variability in judgement, often due to irrelevant factors such as mood and weather \cite{Kahneman2016NOISEHT}. To better illustrate this concept, consider the following example with a zero-sum two vs. two stochastic soccer game played on a five by five grid environment. Both agents of each team initially occupy the same corner of the grid, opposite the agents of the other team. The highest reward comes from moving a ball into the corner originally occupied by the opposing team (“shooting”), with smaller rewards earned for intercepting the ball from the other team, or making a successful pass to an agent’s teammate. The highest cost comes from the opposing team moving the ball into an agent’s original position, with smaller costs incurred for losing the ball to the other team, or making a pass which lands in a peripheral square not occupied by an agent’s teammate. While collecting data on agent performance over ten games, in the case of noisy rationality, during one trial, a “rainy day” may be simulated, such that all agents consistently make incomplete passes, lose the ball to the other team, and fail to shoot the ball. Likewise, “fatigue” could be simulated for individual agents during different trials, causing their performance to be worse. Both of these instances would induce noise in the data. 

Some MIRL methods have modeled noisy rationality, alternatively called Boltzmann-rationality, to account for such unexplained variability within trajectory samples \cite{ziebart2008maximum, Boularias2011RelativeEI}. The most successful method thus far has proven to be the maximum entropy (MaxEnt) IRL framework.

\subsubsection{MaxEnt IRL framework}

MaxEnt IRL assumes observed trajectories are chosen with a probability proportional to the exponent of their value. Assuming the agent is Boltzmann-rational, MaxEnt IRL produces a distribution over behaviors constrained to match feature expectations, while remaining no more committed to a particular trajectory than this constraint requires. The MaxEnt framework thus generates suboptimal, yet plausible, planning algorithms which have been shown to reasonably describe the decision making behavior of humans \cite{ziebart2008maximum, inproceedingsKZ}. Unfortunately, this proof assumes that a reward function is linear in its features, limiting it to use in such cases \cite{ziebart2008maximum}. Regardless, already popular in single-agent IRL methods, the MaxEnt framework can be extended to multi-agent problems, and such extensions show initial success in modeling noisy demonstrations from multiple experts. 

\subsubsection{MaxEnt MIRL for handling noise}

\cite{inga2020inverse} worked to develop theoretical foundations for MaxEnt MIRL through proving the unbiasedness of cost function parameter estimations for three inverse dynamic game methods (with bias referring to the difference between a method’s expected value and the true value of the parameter being estimated). The researchers applied the principle of maximum entropy to identify a probability density function, serving as the basis for cost function parameters in \textit{n}-player games. This allowed them to model the least biased estimate possible for the given information \cite{Jaynes1957InformationTA}. To simulate noisy measurements, the researchers added Gaussian noise to states and controls to induce a specific signal-to-noise ratio (SNR). The method’s capabilities were demonstrated on a simulated two-person ball-on-beam game for multiple solution concepts: a Pareto-efficient solution, open-loop NE and feedback NE \cite{inga2020inverse}. The researchers showed that their method for a Pareto-efficient solution in a nonlinear cooperative dynamic game setting was robust to noisy measurements. However, they found the other methods for competitive dynamics deteriorated when the SNR was less than 20 dB, an effect exacerbated by an increased number of maximum likelihood estimations in the Nash dynamic games. These results indicate that the Pareto-efficient solution for cooperative dynamics, which employs a global cost function representing an aggregation of individual agent cost functions, is more robust to noise than competitive dynamics. This suggests that noise may have different effects either on agent responses in competitive dynamics, or on the ability of MIRL methods to make inferences based on noisy measurements in competitive games. Therefore, further research on noisy measurements in competitive multi-agent games would be useful.

\subsubsection{Limitations of MaxEnt MIRL for noise}

Despite the initially promising results of this work, MaxEnt MIRL requires solving an integral over all possible dynamics while computing its partition function, making it only suitable for small-scale problems \cite{finn2016connection}, such as the two-player dynamics in \cite{inga2020inverse}. Additionally, the concept of noisy rationality alone is insufficient for modeling human behavior due to systematic, non-random deviations from rationality capable of skewing inferences \cite{10.2307/1914185, Evans2015LearningTP}. These deviations consist primarily of biases and heuristics.

\subsection{Biases and heuristics}

Systematic inconsistencies in human decision making frequently manifest as biases, heuristics and the cognitive limitations of learning itself \cite{armstrong2019occams, articleLO}. Biases are disproportionate inclinations to prefer one alternative over another, which may stem from mental qualities including false beliefs and inconsistent judgments of time. Specific biases include loss aversion (the subjective weight assigned to potential losses is larger than that assigned to potential gains), risk aversion (opting for the preservation of capital over the potential for a higher-than-average return), and ambiguity aversion (the tendency to favor the known over the unknown) \cite{10.2307/1914185, 10.2307/1907921, 10.2307/1884324}. Heuristics are decision making shortcuts which result from cognitive limitations including those for reasoning time, memory capacity, and information. EUT also implies logical omniscience, complete knowledge of all logical proceedings from one’s current commitments in combination with any set of choice options. Although psychologically unrealistic, the pillar of logical omniscience in EUT is technically difficult to avoid \cite{10.2307/20116982}. 

In the example of the soccer game, heuristics and biases may influence agent performance in more consistent ways than noise. For example, if interception by the opposing team becomes more likely the longer an agent has possession of the ball, an agent may rely on a heuristic in which they default to always choosing the first viable option that comes to mind as opposed to computing the optimal next move, based on the assumption that good options are generated faster than bad ones. This may lead an agent to choose a worse move (i.e. trying to shoot although the ball may be intercepted by the other team, as opposed to making a pass to their teammate in a better position to do so) because it satisfied their heuristic. In another rendition, an agent may have a distance bias, such that they put more weight on options which are physically closer in space. This bias may also cause an agent to shoot when a pass to a more physically distant teammate may have again been more advantageous. As opposed to noise, which is generally caused by cognitively irrelevant or transitory factors, heuristics and biases are often embedded in an agent’s reasoning processes and will systematically cause decision making to deviate from ideal perfect rationality through time. 

The recognition of these cognitive qualities as well as the limitations of EUT have inspired alternative decision making theories including prospect theory, and regret theory, rank-dependent expected utility \cite{10.2307/1914185, RePEc:inm:oropre:v:30:y:1982:i:5:p:961-981, RePEc:eee:jeborg:v:3:y:1982:i:4:p:323-343} as well as alternative formulations of human rationality \cite{CH2020RF}. None of these alternative formulations have yet taken on the primary role of modeling agent behavior in IRL or MIRL methods, but mechanisms for handling biases and heuristics have shown preliminary success.

\subsubsection{Examples in single-agent IRL}

A few methods in single-agent IRL have addressed handling specific biases. Myopic (short-sighted) planning and hyperbolic time-discounting have been identified via Bayesian inference while an agent’s preferences are simultaneously inferred \cite{10.5555/3015812.3015860, Evans2015LearningTP}. Specific levels of planning depth and impulsivity in an opponent have also been successfully inferred in simple economic games \cite{armstrong2019occams, articleHAM}. These results addressing biases and heuristics are promising for single-agent environments, and relevant intuitions may be applied to multi-agent domains, where deviations from perfect rationality may become more severe.

\subsubsection{Complications for multi-agent environments}

Unfortunately for the tractability of MIRL problems, the more complicated a game dynamic becomes, the more likely it is for strategies to divert from optimality \cite{wang2018competitive}. Game complexity is based on game configuration and state space, with more complex games having larger rule sets and more detailed game mechanics \cite{Allis1994SearchingFS, Heule2007SolvingGD}. Solvability of a game is related to state-space complexity (the number of possible legal game positions) and game-tree complexity (the number of leaf nodes in the solution search tree), both based on the initial game state \cite{Heule2007SolvingGD, 10.1007/978-3-030-01713-2_22}. Based on these definitions, multi-agent games naturally increase game complexity and make identification of a solution more complicated. Such scenarios are prone to further deviations from the assumption of optimality due to amplifying effects of discrepancies from optimality resulting from the decision making processes of multiple potentially heterogeneous agents. Despite the increased complexity, preliminary attempts to handle biases and heuristics in multi-agent scenarios have been developed.

\subsubsection{MaxEnt MIRL for handling biases and heuristics}

\cite{articleNCO} extend the MaxEnt framework to inverse optimal control games to infer reward functions for suboptimal demonstrations in multi-agent interactive scenarios. They model the presence of biases and heuristics in agents through stochastic environmental transition functions by which agents may choose suboptimal actions. In parallel, human decision makers employ deliberate randomization, in which agents deliberately select stochastic choices following a specific preference originating either from the desire to minimize regret, incomplete preferences, difficulty in assessing one’s risk aversion, or other forms of non-expected utility \cite{agranov_stochastic_2017}. The researchers test their method on a series of two-player games, including scenarios where two agents adhere to a leader-follower dynamic. The researchers find that their method is able to reason about the type of behavior caused by interaction rewards and infer reward parameters close to ground truth, even when agents perform with suboptimal policies. 

Similarly, \cite{10.5555/2615731.2615762} consider the case of stochastic transition functions in robotics and employ an extension of MaxEnt IRL, demonstrating success in obtaining a realistic reward function even when the transition function is both stochastic and unknown. While MaxEnt MIRL is useful for addressing deliberate randomization in demonstrations, other methods may be able to capture specific biases and heuristics in more detail. 

\subsubsection{Handling specific biases and heuristics}

\cite{tian2021learning} find that reasoning about an agent’s intelligence level using Theory of Mind (ToM) can capture bounded intelligence and enhance flexibility of MIRL methods in multi-agent settings. ToM is an aspect of human social cognition, referring to our ability to explain and predict other's behaviour by attributing it to independent mental states, including beliefs and desires, supported by neuroscientific evidence \cite{GALLAGHER200377}. Multi-agent settings require interaction models to characterize mutual influence among agents, reasoning which motivated the choice to augment agents with ToM. Modeled as a nested, recursive cognitive reasoning structure and in ToM intelligence is equated to the depth of recursion. Starting from myopic agents with the lowest intelligence level, recursive reasoning is represented in an iterative fashion and an MIRL method reasons about other agents’ intelligence levels during learning. To model irrationality, the researchers use a rationality coefficient to control the degree of agents conforming to optimal behaviors. They validate their approach on zero-sum and general-sum games with synthetic agents in addition to real driving data from a forced merging scenario. Through leveraging this recursive reasoning structure, their approach, termed Cognition-Aware MIRL (CA-MIRL), is able to relax assumptions about an equilibrium solution concept. Additionally, the method is able to jointly learn reward functions in multi-agent games without assuming specific roles for the agents. This is helpful for more accurately representing heterogeneous human behavior. One limitation of this approach is the assumption that intelligence levels of agents are constant. The analysis of driving data used to validate the method suggests that initial periods of competition for dominance in an interaction produce fluctuating intelligence levels. The researchers state that their method extends to \textit{n}-player games, however, they only demonstrate ability on two-player scenarios to compare with pairwise baselines. 

Alternatively to the ToM approach, \cite{wang2018competitive} propose a two-step strategy for handling sub-optimality. In the initial policy step, an adversarial training algorithm solves for a NE strategy in a zero-sum stochastic game parameterized by the current reward function. Agents are pit against their optimal opponent, and an actor-critic proximal policy optimization (PPO) is performed to compute a policy gradient, and subsequently improve policies. To calculate the reward gradient, a batch of sixty-four observations is collected from a set of demonstrations, with each observation providing a gradient calculated in the MIRL algorithm. The incumbent reward function, modeled with a deep neural network, is then updated via this stochastic gradient descent to minimize the performance gap with the optimally-defined reward function. The researchers make the assumption that agents are still performing decently well and the algorithm always considers and relies on both agents to solve for the reward function. The method is shown to be robust to variations in quality of expert demonstrations and is tested on a two vs. two predator-prey game on a five by five grid environment. Compared to other benchmark algorithms for solving competitive MIRL tasks, including Bayesian MIRL and decentralized MIRL, their method is more robust to variation in expert trajectories and can scale to larger games. 

Both of the methods proposed by \cite{tian2021learning} and \cite{wang2018competitive} intentionally do not decouple the agents from one another. Decoupling of reward function inference into agent-level subproblems is common when using a NE solution concept, which thus inherently assumes rational decision making by individual agents. 

\subsection{Future directions in suboptimal planning}

Based on the results of the research on handling noise, biases and heuristics in MIRL, it is clear that multi-agent extension of MaxEnt IRL may be adequate to handle noise and generalized biases and heuristics. More accurate representations may be obtained through modeling specific biases and heuristics, however, incorrect modeling of this type may also lead to biased inference if these qualities do not remain constant. It is worth exploring if MIRL methods may be able to continuously adapt to specific forms of suboptimality in experts. Accommodating these qualities on an individual level may have significant implications for more realistic modeling in dynamic multi-agent interactions, the focus of the next section. 

\section{Multi-agent interactions}

It is difficult to define an optimal policy for expert behaviors in a multi-agent environment because the behaviors of each other agent must be taken into account. This becomes increasingly difficult as the number of agents or search space size increases, or as given environments become more complicated \cite{9308491}. When multi-agent interactions are formalized as games, a central approach for MIRL problems, challenges arise related to the MDPs used to model behavior \cite{Hu2003NashQF}. Primarily, the concept of optimality must be replaced with an equilibrium solution concept, which can be used to model different multi-agent dynamics including two-player zero-sum games, non-zero-sum stochastic games, and cooperative games in which coalitions of agents act independently to achieve a common goal \cite{Oda2021EquilibriumIR}. 

Human actors and their interactions are commonly modeled with formalisms grounded in economic analysis, primarily game theoretic strategies and solution concepts. Dynamic game theory provides a useful mathematical framework for describing the behavior and decision making of multiple agents repeatedly interacting with each other over time, and has successfully been applied in domains outside of economics, including biology \cite{MOLLOY2018754}.

Previous research in MIRL has leveraged models of equilibrium solutions to Markov games to represent social behavior \cite{yu2019multiagent, Gruver2020MultiagentAI}. Different solution concepts make different assumptions about qualities of both agents and game dynamics, and affect the context in which rewards are inferred, making them an influential component of realistic behavioral modeling.

\subsection{The Nash equilibrium}

A simple solution concept commonly used in MIRL is the NE, a stable state in which no agent can gain by a unilateral change of strategy \cite{Reddy2012InverseRL}. It captures the idea that agents ought to perform as well as they can given the strategies selected by other agents, and is proven to exist in all finite and infinite games \cite{Levin2006LearningIG, nash1950equilibrium, Debreu886}. A NE solution concept is commonly used as a framework to model optimal policies which depend on another agent’s policy, and against which to compare observed demonstrations in MIRL research \cite{Reddy2012InverseRL, song2018multiagent}. Using a tightly constrained formalism such as the NE, researchers can gain analytical tractability over strategies and game dynamics. Games modeled by a NE can be decomposed into a sequence of single-stage games with an equilibrium solution concept at each point.

\subsubsection{Limitations of the Nash equilibrium}

Early methods of MIRL relied on a NE solution concept \cite{Reddy2012InverseRL, Lin2018MultiagentIR, articleLIN, song2018multiagent}. However, traditional epistemic characterizations require mutual knowledge beyond beliefs between agents, thus causing this result to fail if agents are mistaken about the strategy choices of others \cite{10.2307/j.ctt7s97x, Bruin2010ExplainingGT}. From an epistemic perspective, therefore, with interest in the strategic reasoning about other’s beliefs and behaviors, an NE solution concept may be of less interest, although it may still prove useful as a solution concept to alleviate strategic uncertainty \cite{Perea2012EpistemicGT, inbookBA}. 
	
It has also been identified that strategies for more complicated games are more likely to divert from optimality, and that a NE, assuming perfect optimality, may lack the ability to handle imperfectly rational agents \cite{wang2018competitive, yu2019multiagent}. Indeed, it has been shown that human behavior deviates from equilibrium solution concepts in real life \cite{Coricelli9163, Selten1998FeaturesOE}. Thus, basic game theoretical formalisms may be less useful for assessing complex, long-term, or abstract trade-offs \cite{GOWDY2008632}.

Particularly, to compute a best response strategy in a game with a NE solution concept, unrealistic assumptions about computational capacity and ability must be made. To find a NE in a traditional representation of a game, all possible payoffs must be explicitly listed, subject to exponential blowup as the number of agents increases. Results from complexity theory demonstrate the excessive cognitive demands a NE strategy places on agents. \cite{SCHGRV} examine various games to determine if a given strategy is a NE, to find a NE, to determine if there exists a pure NE, and to determine if there exists a NE in which payoffs to a player meet certain guarantees. The researchers show that all problems are complete for various complexity classes. \cite{1366245} also shows that for a class of bimatrix games, computing a NE, even in the best case, requires exponential time.

As previously stated, humans exhibit bounded intelligence and deviate from rationality in many ways \cite{Coricelli9163, Goeree2000TenLT, 10.2307/4502047}. It is believed that many of these cognitive adaptations are a result of the more complex systems of the human world, and ignorance to them could fundamentally change social interactions. Indeed, animals of lower cognitive levels tend to behave much closer to a rational actor model, such as ducks that will split into a stable NE when presented with two different sources of intermittently dispersed food delivering different expected payoffs \cite{GOWDY2008632}. Treating such qualities as anomalies may reduce their centrality to an authentic human decision making process. 

Human decision making also cannot be accurately predicted without reference to social context \cite{GOWDY2008632}. A recently presented concept, hyper-rational choice, describes a process in which an agent accounts for the gains and losses of other agents in addition to their own, before choosing an action. Hyper-rational choice can help to model the behavior of people considering environmental conditions, valuation systems of individual agents, and systems of societal beliefs and values \cite{askariBehavioralModelGame2019}. Due to the necessity of accounting for the behavior of all other agents, a NE also becomes harder to apply as the number of agents or size of search space increases \cite{9308491}. \cite{articleDAS} shows that identifying a NE in three-player games is PPAD-complete, a subclass of TFNP. If the NE sought requires special properties (i.e. optimizing total utility), the problem typically becomes NP-complete \cite{GILBOA198980, CONITZER2008621}. Determining whether a game has a pure NE is also NP-hard \cite{articleGOT}. 

NE solution concepts also yield further challenges. For example, agents’ strategies are often mixed and cannot be inferred exactly from finite observations of actions in each state. Thus, one cannot model a strategy as simply an observation as can be done in IRL. Strategies must instead be treated as latent variables that are not directly observed, bridging the gap between reward function and observations. It is also common for a NE to be non-unique, which presents the challenge of reward unidentifiability, in which multiple reasonable solutions to an inversion model may exist and multiple inversion models may be equally sensible to solve a problem.

These limiting factors of the NE solution concept motivate the use of alternative solution concepts in recent MIRL research. 

\subsection{Alternative solution concepts}

To address the shortcomings of NE approaches, many alternative solution concepts to the NE are now being explored in MIRL. The most successful approaches are more generalized forms of the NE which allow for less strict behavioral assumptions. Additionally, methods which employ recursive updating of policies based on identified cooperative trajectories show successful results, at least in applications to cooperative dynamics. 

\subsubsection{Feedback Nash equilibrium}

A feedback NE employs a different information pattern from the traditional open-loop NE formalism, allowing it to more closely approximate natural decision making processes. Information patterns (i.e. open-loop or closed-loop patterns) are structured patterns for interacting with and organizing information. Different patterns are associated with different dynamic games, and result in fundamentally different solutions \cite{Baar1998DynamicNG}. In an open-loop formation, the simplest solution, the game’s repeated nature is ignored and time is the only information an agent takes into account at each stage. Excluding private and global state variables (which are used to track game history in closed-loop formations) from the inclusions used to compute optimal agent strategies makes open-loop formations more tractable \cite{wiszniewska-matyszkiel_open_2014}. Game stages can be combined into a single static game and the entire trajectory chosen at once to satisfy equilibria. The resultant static games thus permit tractable methods for analysis and computing \cite{facchineiGeneralizedNashEquilibrium2007, facchineiGeneralizedNashEquilibrium2009}. 

However, dynamic agent nature is ignored, representational expressiveness of solutions is limited, and agent behavior ends up diverging from the equilibrium which would emerge with dynamic modeling. Intelligent game play in repeated games involves observing the evolving game and responding accordingly. This behavior can more accurately be modeled through a closed-loop formation, the basis of a feedback NE. In a closed-loop information pattern, agents choose control policies which define input as a function of the state at specific stages. Policies at each stage constitute equilibria for dynamic subgames over subsequent game stages. Therefore, capturing agent strategies which anticipate and respond to the reactions of other agents becomes possible \cite{laine2020multihypothesis}. Numerical routines for computing a feedback NE (beyond simple cases or cases with strong constraints on state and input dimensions) are challenging, yet efficient solution methods for linear quadratic and nonlinear cases do exist \cite{laine2021computation}. 

\cite{inga2020inverse} examine cooperative games with Pareto-efficient solutions and competitive games with open-loop and feedback NE solution concepts using linear quadratic games. The Pareto-efficient solution can be described by a global cost function given by the sum of uniformly weighted agent cost functions, the parameters of which are identified from trajectories describing a Pareto-efficient solution. Their results show that all methods determine cost function parameters which correctly explain the observed trajectories. 

In a feedback NE, an agent may base its plays on information tracked over time. Therefore, an agent may have the ability to choose a play which is advantageous in consideration of an opponent's biases. For example, in the competitive soccer game, if agents on the opposing team consistently fail to make successful shots on an agent’s own goal, an agent may change its strategy to be more offensive, based on the assumption that there is a decreased need for defensive effort due to past play outcomes. 

\subsubsection{Correlated equilibrium}

Another alternative solution concept is the correlated equilibrium (CE), a superset of NE which describes a joint strategy profile in which no agent can achieve higher expected reward through unilateral policy change \cite{AUMANN197467}. More general than a NE, a CE does not require independent agent interactions in each state. Table 2 provides an example of correlated equilibrium for the following two-player game:

Two agents, competitive drivers approach an intersection from perpendicular streets. Each agent has two strategies, which for both agents are (stop, go). The utilities below reflect the situation in which the strategies of the first agent are rows and of the second agent are columns and the following five distributions reflect correlated equilibria for this game: 

\begin{figure}
    \centering
\begin{tabular}{|l|l|l|lllllllllllllll}
\cline{1-3}
     & stop & go   &                       &                        &                        &                       &                        &                        &                       &                          &                          &                       &                          &                          &                       &                          &                          \\ \cline{1-3} \cline{5-6} \cline{8-9} \cline{11-12} \cline{14-15} \cline{17-18} 
stop & 4, 4 & 1, 5 & \multicolumn{1}{l|}{} & \multicolumn{1}{l|}{0} & \multicolumn{1}{l|}{1} & \multicolumn{1}{l|}{} & \multicolumn{1}{l|}{0} & \multicolumn{1}{l|}{0} & \multicolumn{1}{l|}{} & \multicolumn{1}{l|}{1/4} & \multicolumn{1}{l|}{1/4} & \multicolumn{1}{l|}{} & \multicolumn{1}{l|}{0}   & \multicolumn{1}{l|}{1/2} & \multicolumn{1}{l|}{} & \multicolumn{1}{l|}{1/3} & \multicolumn{1}{l|}{1/3} \\ \cline{1-3} \cline{5-6} \cline{8-9} \cline{11-12} \cline{14-15} \cline{17-18} 
go   & 5, 1 & 0, 0 & \multicolumn{1}{l|}{} & \multicolumn{1}{l|}{0} & \multicolumn{1}{l|}{0} & \multicolumn{1}{l|}{} & \multicolumn{1}{l|}{1} & \multicolumn{1}{l|}{0} & \multicolumn{1}{l|}{} & \multicolumn{1}{l|}{1/4} & \multicolumn{1}{l|}{1/4} & \multicolumn{1}{l|}{} & \multicolumn{1}{l|}{1/2} & \multicolumn{1}{l|}{0}   & \multicolumn{1}{l|}{} & \multicolumn{1}{l|}{1/3} & \multicolumn{1}{l|}{0}   \\ \cline{1-3} \cline{5-6} \cline{8-9} \cline{11-12} \cline{14-15} \cline{17-18} 
\end{tabular}
    \caption*{Table 2: Decision matrix for the driving game showing the payoffs for each agent based on selected strategy and five possible CE for the game.}
    \label{fig:my_label1}
\end{figure}

In a CE, a probability distribution exists over the joint action space and all agents optimize payoff with respect to one another’s probabilities \cite{articleLIN, articleGRN}. One advantage of a CE is that it is less computationally expensive than a NE. For example, a CE can be found in polynomial time via linear programming, whereas NE requires finding its fixed point completely \cite{articlePAPA}. It is possible for two agents to respond to each other’s historical game plays and end up converging to CE, and many decentralized adaptive strategies naturally do so \cite{10.1145/1390156.1390202, https://doi.org/10.1111/1468-0262.00153, 10.2307/2329400}.

\cite{articleLIN} explores both CE and NE within a subset of MIRL problems distinguished by unique solution concepts. In the CE formulation, agents attempt to maximize total game value. A linear programming problem with constraints defines necessary and sufficient conditions for observed policies to be a CE and solutions which minimize the difference between the observed bipolicy and a local cooperative solution are selected. The researchers find that this strategy accurately estimates the value of the true reward function. Interestingly, equivalent accuracy is achieved when following a NE strategy in the same formalism. Both methods achieve exceptional results on two grid-world games, measured via the metric of normalized root mean squared error (NRMSE). Table 3 shows the results for CE-MIRL and NE-MIRL methods compare to the results of other methods tested on the same environment \cite{articlePAPA}.

\begin{figure}
    \centering
\begin{tabular}{|l|l|l|}
\hline Method    & Grid 1   & Grid 2 \\
\hline
\textbf{CE-MIRL} & \textbf{1.30} x $10^{-3}$ & \textbf{1.39} x $10^{-10}$ \\ \hline
IRL     & 0.287       & 0.311        \\ \hline
dMIRL   & 0.089       & 0.083        \\ \hline
\textbf{NE-MIRL} & \textbf{0}           & \textbf{0}            \\ \hline
IRL     & 0.271       & 0.283        \\ \hline
dMIRL   & 0.104       & 0.103        \\ \hline
\end{tabular}
    \caption*{Table 3: NRMSE for CE and NE-MIRL methods compared to IRL and dMIRL methods when tested on two grid world games.}
    \label{fig:my_label2}
\end{figure}

The authors note that this equivalent performance could be due to restricted testing environment conditions producing no noise, generation of the bipolicy from a corresponding MRL-Q learning algorithm, or the incorporation of strong prior game information into the methods. Therefore, further research is needed to compare CE and NE performance.

\subsubsection{Logistic stochastic best response equilibrium}

Quantal response equilibrium (QRE) is a mixed strategy profile such that, for every player, a mixed strategy is the stochastic response to an agent’s adversaries’ mixed strategies. A QRE is a statistical version of a NE, unique when the error variance is sufficiently large such that players’ responses are nearly uniformly distributed, regardless of the competitors’ strategies. This may then result in the uniqueness of a QRE \cite{MCKELVEY19956}.

Logistic quantal response equilibrium (LQRE), a form of the QRE, is a solution concept in which agents are assumed to make errors in choosing which pure strategy to play. The probability of any chosen strategy is positively related to its payoff, computed via a logistic quantal response function based on beliefs about other agents’ probability distributions over strategies. LQRE allows every strategy to be played with non-zero probability such that any trajectory is possible.

\cite{Oda2021EquilibriumIR} develops spatial equilibrium IRL (SEIRL), using a LQRE to model an uncooperative network model for passenger-seeking driver behavior. A Markov potential game based on a transportation network, agents compete for a common resource (passengers) while seeking to minimize individual cost while choosing the best route to a destination. Equilibrium value iteration is performed to derive an optimal policy for multiple vehicles in equilibrium, improving imitation accuracy. Table 4 shows the performance in mismatched distance ratio of SEIRL compared to other benchmark methods. The experiments were conducted with full access to data as well as for two conditions with a missing percentage of data set to either five or ten percent, results for which are shown in Table 4 \cite{Oda2021EquilibriumIR}.  

\begin{figure}
    \centering
\begin{tabular}{l|llllll|}
\cline{2-7}
                                     & \multicolumn{2}{l|}{Full data}                & \multicolumn{2}{l|}{5\% missing}              & \multicolumn{2}{l|}{10\% missing} \\ \hline
\multicolumn{1}{|l|}{Policy}         & Trial 1        & \multicolumn{1}{l|}{Trial 2} & Trial 1        & \multicolumn{1}{l|}{Trial 2} & Trial 1         & Trial 2         \\ \hline
\multicolumn{1}{|l|}{Expert}         & 0.099          & 0.123                        & 0.099          & 0.123                        & 0.099           & 0.123           \\ \hline
\multicolumn{1}{|l|}{\textbf{SEIRL}} & \textbf{0.223} & \textbf{0.328}               & \textbf{0.363} & \textbf{0.485}               & \textbf{0.411}  & \textbf{0.543}  \\ \hline
\multicolumn{1}{|l|}{Opt}            & 2.867          & 2.381                        & 2.867          & 2.381                        & 2.867           & 2.381           \\ \hline
\multicolumn{1}{|l|}{SE-Opt}         & 0.943          & 1.076                        & 0.943          & 1.076                        & 0.943           & 1.076           \\ \hline
\end{tabular}
    \caption*{Table 4: Results in mismatched distance ratio for the SEIRL method on passenger-seeking driver behavior data. Results are compared to Expert behavior as well as two other baselines (Opt, SE-Opt) \cite{Oda2021EquilibriumIR}.}
    \label{fig:my_label3}
\end{figure}

A LQRE can also introduce the concept of stochastic best response (SBR) dynamics. Assuming a LQRE solution concept, \textit{n}-player games with payoffs subject to random error naturally induce SBR dynamics. For example, if \textit{n} populations of agents are randomly and repeatedly matched to play a game, in each period, if one agent revises their strategy to maximize random payoff realization, a stochastic process produces SBR dynamics \cite{Ui2002QuantalRE}. 

Building on this idea, \cite{yu2019multiagent} introduces logistic stochastic best response equilibrium (LSBRE) to rationalize suboptimal demonstrations from computationally bounded agents by maximizing the likelihood of expert trajectories with respect to the stationary LSBRE distribution. In the LSBRE, a strategy profile is defined such that, given what other players are doing, a strategy is the best response if and only if a player cannot gain more utility by switching to a different strategy. LSBRE explicitly defines tractable joint strategy profiles used to maximize likelihood of expert demonstrations and characterizes a trajectory distribution induced by parameterized reward functions \cite{yu2019multiagent}. These trajectories can be characterized by an energy-based formulation in which the model associates an energy value with a sample x, modeling the data as a Boltzmann distribution \cite{inbookLEC}. The energy function parameters are chosen to maximize the likelihood of the data, thus, the probability of a trajectory increases exponentially with the sum of its rewards \cite{inbookLEC}. LSBRE corresponds to the result of repeatedly applying stochastic, entropy-regularized, best response mechanisms while each agent optimizes its own actions. Bridging the optimization of joint likelihood and conditional likelihood with maximum pseudolikelihood estimation uncovers a relationship between MaxEnt IRL and LSBRE. 

The method proposed by \cite{yu2019multiagent}, multi-agent adversarial IRL (MA-AIRL) is tested across different cooperative (navigation and communication) and competitive (keep-away) tasks. MA-AIRL outperforms multi-agent generative adversarial imitation learning (MA-GAIL) benchmarks even when given no prior knowledge of interactions, as described by results in Table 5. The first table shows statistical correlations between learned reward functions and ground-truth rewards for the cooperative tasks with mean and variance taken across \textit{n} independently learned reward functions for \textit{n} agents. The second table shows the statistical correlations between the learned reward functions and the ground-truth reward functions for the competitive task.

\begin{figure}
    \centering
\begin{tabular}{|l|l|ll|llll}
\cline{1-4} \cline{6-8}
Task  & Metric & MA-GAIL & \textbf{MA-AIRL} & \multicolumn{1}{l|}{} & \multicolumn{1}{l|}{Algorithm} & MA-GAIL & \multicolumn{1}{l|}{\textbf{MA-AIRL}} \\ \cline{1-4} \cline{6-8} 
Nav.  & SCC    & 0.792   & \textbf{0.934}   & \multicolumn{1}{l|}{} & \multicolumn{1}{l|}{Av. SCC}   & 0.538   & \multicolumn{1}{l|}{\textbf{0.721}}   \\
      & PCC    & 0.556   & \textbf{0.882}   & \multicolumn{1}{l|}{} & \multicolumn{1}{l|}{Av. PCC}   & 0.445   & \multicolumn{1}{l|}{\textbf{0.694}}   \\ \cline{1-4} \cline{6-8} 
Comm. & SCC    & 0.879   & \textbf{0.936}   &                       &                                &         &                                       \\
      & PCC    & 0.612   & \textbf{0.848}   &                       &                                &         &                                       \\ \cline{1-4}
\end{tabular}
    \caption*{Table 5: Result comparison between MA-GAIL and MA-AIRL methods in Spearman's rank correlation coefficient (SCC) and Pearson's correlation coefficient (PCC) \cite{yu2019multiagent}. }
    \label{fig:my_label4}
\end{figure}

\cite{ALSMARG} further test an LSBRE-based method on data from three shared space locations in Vancouver and New York City, demonstrating success with suboptimal demonstrations, mixed-strategy policies, and social negotiation in road user interactions. Their method uses the LSBRE solution concept to effectively coordinate multi-agent decisions. However, the authors note that behavioral and environmental norms may not translate to other situations, and recommend future work across different cultural norms.

\subsubsection{Entropy regularized mean field Nash equilibrium}

Mean field games (MFGs) provide a method for modeling populations as a whole and can be solved via a mean field NE (MFNE). Insights from research on population modeling can inform the validity of using a MFG to represent a population. MFGs stem from a branch of game theory which provides tractable models for large agent populations through considering the limit of \textit{n}-player games as \textit{n} approaches infinity. The agent population is represented as a distribution over a state space and an agent’s optimal strategy is informed by a reward, a function of population distribution and aggregate actions. This choice of representation is attractive as it is scalable to arbitrary population sizes by representing population density as opposed to individual agents. However, the reduction of a MFG to a multi-agent MDP is impossible unless every agent is modeled individually. 

A tacit assumption exists that for high numbers of agents, agent-level models can be replaced by computationally advantageous population-level models. However, agent-level models can have the advantage of being more realistic. \cite{articleBOS} examines the viability of replacing an agent-level model with a population-level model when examining the influence of the global economy on the risk level of individual agents’ investment decisions. The agent based model assumes \textit{n} heterogeneous agents, with different risk levels. In the population model, only the average risk level is considered. The study finds that for larger populations, replacement with a population-level model held, whereas it did not for smaller populations. However, the researchers find the opposite result when modeling the spread of an epidemic, suggesting appropriate model choice depends on the target dynamic. The study was limited to agent-based simulations with homogeneous models employing the same parameters, differing only in value. Thus, different results may occur with heterogeneous agent models \cite{articleBOS}. 

Despite the necessity of determining whether a population-level formulation is appropriate for a particular scenario, researchers have explored using MFGs to model populations in MIRL problems. Computational intensity and algorithmic run-time for computing traditional NE solution concepts grows exponentially with expansion of joint state-action spaces when population size increases \cite{Hu2003NashQF, BOWLING2002215, articleDAS}. A MFNE solution concept simplifies this computation for a population model through considering its asymptotic limit, assuming that agents in the population are homogeneous, and that the population approaches infinity \cite{chen2021agentlevel}. Mean-field approximations leverage the empirical distribution representing aggregated population behaviors, reducing interactions to a dual-view interplay of a single agent and the whole population. Unfortunately, much like NE, MFNE only holds for socially optimal situations. Thus, this method suffers from biased inference and additional errors if expert demonstrations are sampled from an ordinary MFNE, as multiple equilibria may exist \cite{bardi2018nonuniqueness, dianetti2019submodular, DELARUE20201000, 10.5555/3306127.3331700}. 

To resolve this issue, by incorporating causal entropy regularization into rewards, \cite{chen2021agentlevel} extends MFNE to the entropy-regularized MFNE (ERMFNE) achieving uniqueness and the ability to accommodate suboptimal demonstrations by agents. ERMFNE serves as a generalization of MaxEnt IRL to MFGs. The researchers show that this method is effectively able to recover ground truth rewards for MFGs \cite{chen2021agentlevel}.

\subsubsection{Archive multi-agent adversarial IRL}

\cite{9308491} propose archive multi-agent adversarial IRL (AMAIRL), formalized as a MaxEnt IRL extension. This method achieves a unique solution through archiving (via storage in shared memory) cooperative trajectories found in inner-loop learning, and replacing original expert trajectories with archived cooperative trajectories intermittently.

\begin{figure}
    \centering
\begin{tabular}{l|l|l}
\cline{1-1} \cline{3-3}
\multicolumn{1}{|l|}{Input: $Expert^1$}                                                                                                                                   &                                                                                        & \multicolumn{1}{l|}{Input: $Experts^N$}                                                                                                                                  \\ \hline
                                                                                                                                                                        & Environment                                                                            &                                                                                                                                                                        \\ \hline
\multicolumn{1}{|l|}{$Agent^1: s, a, r$}                                                                                                                                  &                                                                                        & \multicolumn{1}{l|}{$Agent^N: s, a, r$}                                                                                                                                  \\ \hline
                                                                                                                                                                        & \begin{tabular}[c]{@{}l@{}}{[}Shared Memory{]}\\ Cooperative Trajectories\end{tabular} &                                                                                                                                                                        \\ \hline
\multicolumn{1}{|l|}{\begin{tabular}[c]{@{}l@{}}{[}Memory{]}\\ $Q-value^1$\\ $Reward Function^1$\\ $Individual Trajectories^1$\end{tabular}}                                  &                                                                                        & \multicolumn{1}{l|}{\begin{tabular}[c]{@{}l@{}}{[}Memory{]}\\ $Q-value^N$\\ $Reward Function^N$\\ $Individual Trajectories^N$\end{tabular}}                                  \\ \cline{1-1} \cline{3-3} 
\multicolumn{1}{|l|}{\begin{tabular}[c]{@{}l@{}}{[}Mechanism{]}\\ MaxEnt IRL\\ Cooperative Archive\\ Individual Archive\\ Expert Trajectory\\ Replacement\end{tabular}} &                                                                                        & \multicolumn{1}{l|}{\begin{tabular}[c]{@{}l@{}}{[}Mechanism{]}\\ MaxEnt IRL\\ Cooperative Archive\\ Individual Archive\\ Expert Trajectory\\ Replacement\end{tabular}} \\ \cline{1-1} \cline{3-3} 
\multicolumn{1}{|l|}{Output: $Reward Function^1$}                                                                                                                         &                                                                                        & \multicolumn{1}{l|}{Output: $Reward Function^N$}                                                                                                                         \\ \cline{1-1} \cline{3-3} 
\end{tabular}
    \caption*{Table 6: Memory storage architecture used by \cite{9308491} Expert input goes to each respective agent (below), and cooperative trajectories in shared memory are archived from and acted out on the environment.  }
    \label{fig:my_label5}
\end{figure}

In this scheme, a cooperative trajectory is any combination of trajectories in which all agents do not interfere with each other while navigating a maze problem. The other agents start to learn cooperative behaviors according to the original selfish expert behaviors, and continue according to the updated expert behaviors. When compared to MaxEnt IRL, AMAIRL is able to find cooperative behavior within a significantly fewer number of steps. Unfortunately, AMAIRL requires a long period of observation to archive and update cooperative behaviors, and computational costs increase as cooperative behaviors deviate further from original expert trajectories.

\subsection{Future directions in alternative solution concepts}

The results of the above research show success for alternative solution concepts, and suggest two main strategies to achieve solutions capable of accommodating natural human behavior. The first group of solution concepts can be characterized by a quality recursive updating, including the feedback NE and AMAIRL. The second group can be characterized by relaxing some of the assumptions which restrict traditional NE to optimal demonstrations. This group includes the CE, LSBRE, and the ERMFNE. A solution concept which employs both of these characteristics may prove useful, as well as concepts which incorporate insights about specific deviations from optimality.

\section{Conclusion}

This research aimed to analyze and compare strategies for handling noise, biases and heuristics in MIRL, as well as alternative solution concepts which accommodate suboptimal decision making and realistic multi-agent interactions. It was found that extensions of the MaxEnt framework to multi-agent scenarios can accommodate suboptimal decision making, and that modeling ToM in agent reasoning can lead to more flexible methods. Additionally this research found that generalizations of NE solution concepts (CE, LSBRE, and ERMFNE) as well as recursive approaches (feedback NE and AMAIRL) are well suited to handle suboptimal agent dynamics. 

Future work in this area may include research into the benefits of modeling specific biases and heuristics over more general approaches, as well as more closely comparing the performance of these alternative solution concepts in more complicated environments. Additionally, in research which tests methods on real-world data, using data which represents a variety of social norms would be useful.

\section*{Acknowledgements}

I would like to thank Tamay Besiroglu for his mentorship throughout this research; I would also like to thank Lawrence Chan (Center for Human Compatible AI, UC Berkeley) for his helpful feedback. This research was supported by the Stanford Existential Risk Initiative (SERI) and the Berkeley Existential Risk Initiative (BERI).

\printbibliography
\end{document}